# A Comparative Analysis of Retrieval Techniques in Content Based Image Retrieval

Mohini. P. Sardey[1], G. K. Kharate[2]

[1] AISSMS Institute Of Information Technology, Savitribai Phule Pune University, Pune (Maharashtra State), India.
sardeymp@yahoo.com

[2] Matoshri COE and Research Center, Savitribai Phule Pune University, Pune (Maharashtra State), India.
gkkharate@gmail.com

***Abstract:*** *Basic group of visual techniques such as color, shape, texture are used in Content Based Image Retrievals (CBIR) to retrieve query image or sub region of image to find similar images in image database. To improve query result, relevance feedback is used many times in CBIR to help user to express their preference and improve query results. In this paper, a new approach for image retrieval is proposed which is based on the features such as Color Histogram, Eigen Values and Match Point. Images from various types of database are first identified by using edge detection techniques .Once the image is identified, then the image is searched in the particular database, then all related images are displayed. This will save the retrieval time. Further to retrieve the precise query image, any of the three techniques are used and comparison is done w.r.t. average retrieval time. Eigen value technique found to be the best as compared with other two techniques.*

## 1. Introduction

Key task of the computer science is always been the management of digital information. Years ago the data used to be in terms of numbers and [1] [4] [9] text; relational databases handled the storage and process of searching. But with the rapid growth of complex data types e. g. images, sound, videos, there is a need to research that suit our changing needs better. So in last two decades, a new approach data management has been studied extensively to address these requirements. In variety of application areas based on content based or on similarity searching has become fundamental computational task [18] which includes multimedia information retrieval, data mining, pattern recognition, data compression, biomedical databases, statistical data analysis. Rest of the paper is organized as follows. Section 2 illustrates frame work of CBIR, section 3 explains various techniques used to implement the objective, section 4 will tabulate the result and finally section 5 draws the conclusion remark and future scope is discuss.

## 2. CBIR Framework

Retrieval system proposed in this paper is described by the frame work or the block diagram as depicted in figure 1. This block diagram explains the flow of the proposed technique very easily which every reader can understand.

Image retrieval system can be conceptually described by the framework depicted in figure 1. In this article we survey how the user can formulate a query, which is the appropriate retrieval technique for various types of image database such as Face, Vehicle, Animal and Flower, how the matching can be done [4].

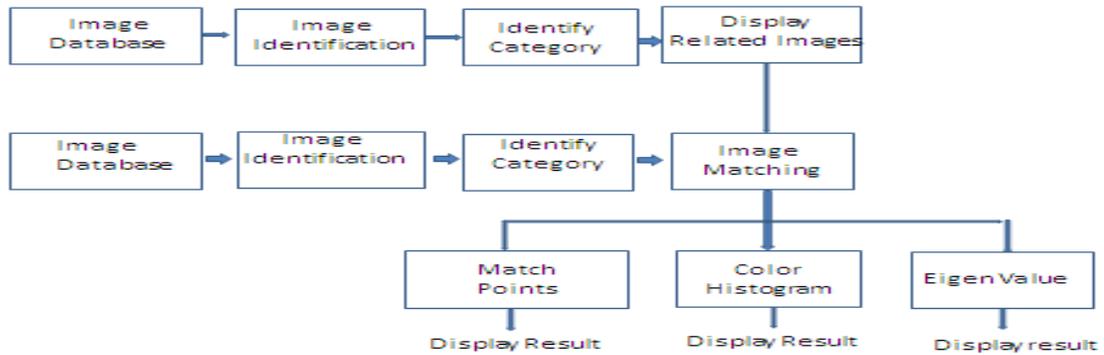

**Figure 1. Basic block diagram of CBIR**

This paper proposes a technique of image retrieval which first identifies the type of image by using edge detection technique, as shown in fig.2. This step is essential when the images are from different format. When the type of query image is known, then system will search the query image in that particular data type only which will save search time substantially. Query image will be searched precisely by using either of the three image retrieval techniques, Color Histogram, Eigen values base and Match point base.

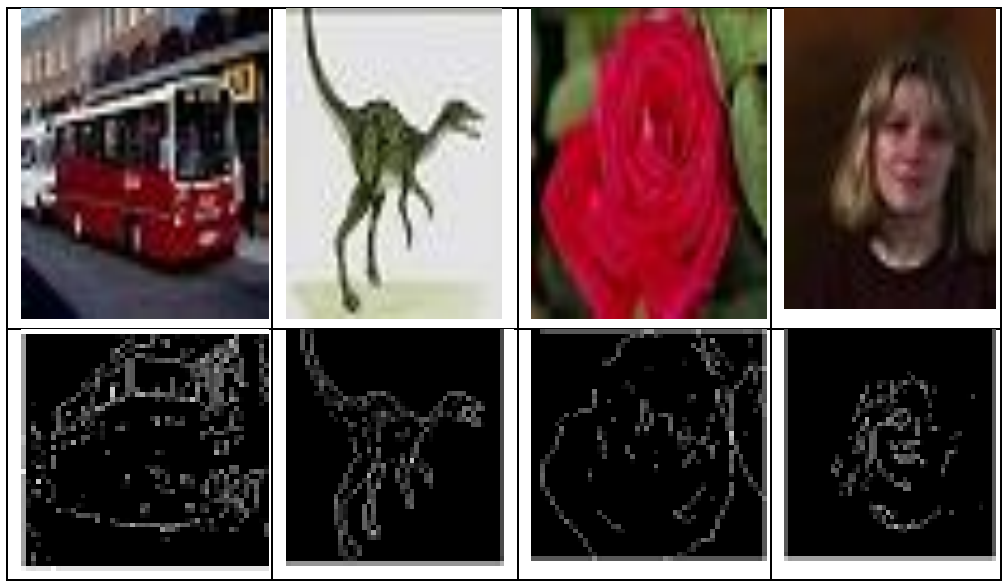

**Figure 2. Image and its Edge**

# 3 Implementation Techniques:

**3.1 Eigen Values:** The term "eigenvalue" is a partial translation of the German "Eigen wert". A complete translation would be something like "own value" or "characteristic value," but these are rarely used. Eigenvalues play an important role in situations where the matrix is a transformation from one vector space onto itself.

When matrix transformation is from one vector space, role played by Eigen value is very important. When applications are based on image processing, eigen value approach plays a prominent role, e.g. measurement of sharpness of an image or segregation of images into images of vehicles or animals, etc. , Aim is to implement the mode with some real time variation, to precise face or image and retrieve it from a large number of stored faces. The Eigen face approach uses Principal Component Analysis (PCA) algorithm for the recognition of the images. It gives us efficient way to find the lower dimensional space.

### 3.1.1 Sensitivity and accuracy of Eigen value:

Basically, eigen value is a matrix which is susceptible to any deviation or changes i.e. disorder in matrix element will result in significant changes in eigen values. When the operations are related to floating point arithmetic, computations will result in introduction of round-off errors and also have similar effect to the perturbations taking place in original matrix [9]. This will in turn result in the magnification of round off errors in the eigen values that are computed.

Assuming A has full set of linearly independent eigenvectors and using the eigenvalue decomposition we can get a rough idea of the sensitivity.

Equation of eigen value and eigen vector for a square matrix can be written as

$$(A-\lambda I) x = 0, x \neq 0$$

This implies that $(A-\lambda I)$ is singular and hence

$$det(A. ëI) = 0$$

This particular definition of eigen value, which excludes the corresponding eigen vector [10], is the characteristic polynomial of A or the characteristic equation and the degree of this polynomial is the order of matrix. Therefore if there are n-eigenvalues, matrix is of size n-by-n.

| Techniques  Database | Match Point (sec) | Histogram | Eigen values |
|---|---|---|---|
| Flower | 7.701922(184x2) | 12.703803 | 3.121432 |
| Face | 6.544435(187x2) | 11.324701 | 3.008408 |
| Vehicle | 6.083464(188x2) | 12.393825 | 3.005223 |
| Animal | 5.851284(185x2) | 11.062617 | 3.678530 |

**Table 1. Retrieval time in each image category**

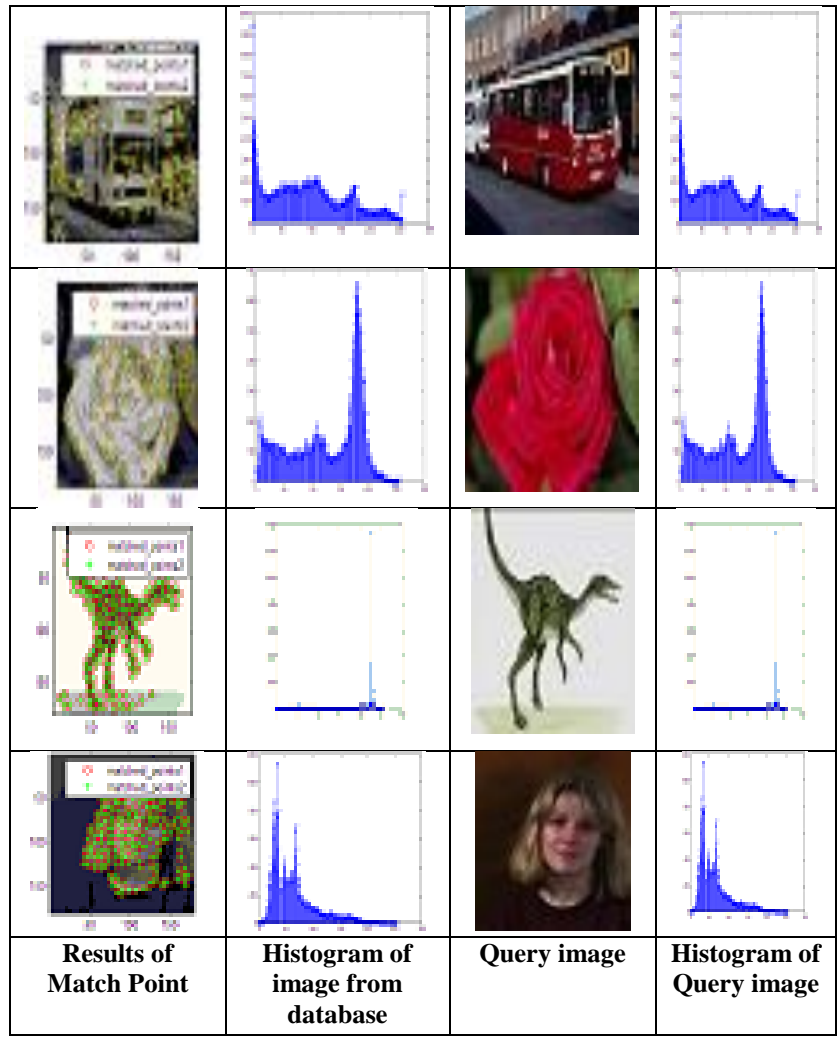

**Figure 3. Results of images in each category**

**3.2 Color :** Color is an important visual attribute for both human perception and computer vision and one of the most widely used visual features in image retrieval [1]. But an appropriate color space and color quantization must be specified along with a histogram representation of an image for retrieval purpose. Histogram describes the global distribution of pixels of an image [17][18]. Main advantage of a color histogram is its small sensitivity to variations in scale, rotation and translation of an image. We utilize different kinds of quantization schemes for the implementation of the color histograms in HSV color space. We observed that the HSV color model is better than the RGB color model for our approach using the following quantization scheme where each color component is uniformly quantized. . Although the color-based methods perform surprisingly well, [14] [15] their performance is still limited to less than 50% in precision. The main reason is because the color representation is low-level, even with the use of pseudo object models.

In general, color is one of the most dominant and distinguishable low-level visual features in describing image. Many CBIR systems employ color to retrieve images, such as QBIC system and Visual SEEK.

The retrieval method of using color characteristic was originally proposed by Swain and Ballard, they put forward the color histogram [6] method of which the core idea is to use a certain color space quantization method for color quantization, and then do statistics for the proportion of each quantitative channel in the whole image color. Abscissa represents the normalized color value, ordinate represents the sum of image pixels which corresponding to each color range [8][9][10]. Image statistical histogram is a one-dimensional discrete function:

$$h_k = n/n_k , \ k=0,1, \ldots,L-1$$

The letter *k* presents eigenvalues of color, letter *l* presents the number of features of value . So we get the color histogram of the image P as follows:

$$H(p)=[h_1, \ h_2, \ \ldots h_{L-1}]$$

There are many color histogram methods such as the global color histogram, cumulative histogram and sub-block histogram. However, color histogram has its own drawbacks, such as the color histograms of different images may be the same.

### 3.3 Match Point Based: Computer Vision System Toolbox is used for this feature:

This paper uses the functions from Computer Vision System toolbox to detect the objects using the Viola-Jones algorithm. Detection of corners in a grey scale image another function is used. Another function is used to detect the corners in a grey scale image. It returns location as a matrix of [x, y] coordinates. The object finds corner in an image using Harris corner detection, minimum Eigenvalues or local intensity comparison method. Using another function, feature vectors are extracted from intensity or binary image. These vectors are also known as descriptors and are derived from pixels surrounding an interest point by the function. These pixels match features and describe them by a single-point location specification. The function extracts feature vectors from an input intensity or binary image. These feature vectors, also known as descriptors are returned as M-by-N matrix having M feature vectors and each descriptor having length N. Corresponding to each descriptor, M number of valid points is also returned. To match the features, match features function is used. To display corresponding feature points an overlay of pair of images in addition to a color-coded plot of corresponding points connected by a line, but the location is defined in the Surf point objects. [21]

## 4 Comparative analysis of all the three techniques:

Following table gives comparison of retrieval time of all types of databases with three different techniques, such as match Point, Histogram and Eigen Values.

| Techniques / Database | Match Point (sec) | Histogram | Eigen values |
|---|---|---|---|
| Flower | 5.357360 | 11.689926 | 2.743924 |
| Flower | 15.029603 | 12.321363 | 4.988143 |
| Flower | 6.987936 | 13.361342 | 3.311586 |
| Flower | 6.152623 | 12.084985 | 3.162717 |
| Flower | 6.943955 | 26.030384 | 5.088805 |
| Flower | 7.701922 | 12.703803 | 3.121432 |
| Face | 7.113300 | 11.335935 | 2.323375 |
| Face | 6.544435 | 11.324701 | 3.008408 |
| Face | 5.480784 | 17.273292 | 3.655586 |
| Face | 6.458972 | 12.654557 | 2.164110 |
| Face | 6.9511341 | 11.463735 | 4.656659 |
| Face | 7.245466 | 11.218740 | 4.571890 |
| Face | 6.291519 | 13.056973 | 1.921271 |
| Face | 5.202477 | 12.512681 | 2.388528 |
| Face | 6.159422 | 12.477321 | 8.896353 |
| Vehicle | 5.205715 | 11.218377 | 2.711211 |
| Vehicle | 6.210938 | 11.659295 | 3.566008 |
| Vehicle | 8.832045 | 11.472673 | 3.346753 |
| Vehicle | 7.789522 | 13.174998 | 2.197203 |
| Vehicle | 9.015264 | 12.895586 | 5.521553 |
| Vehicle | 6.083464 | 12.393825 | 3.005223 |
| Vehicle | 9.241839 | 16.0635481 | 3.049684 |
| Vehicle | 5.151585 | 11.022538 | 2.063009 |
| Vehicle | 5.835444 | 17.895952 | 2.903603 |
| Vehicle | 5.636356 | 10.383492 | 2.820357 |
| Animal | 5.851284 | 11.062617 | 3.678530 |
| Animal | 4.050715 | 11.664695 | 4.105603 |
| Animal | 6.812625 | 11.303356 | 2.770873 |
| Animal | 6.196215 | 10.947030 | 6.019186 |
| Animal | 5.685910 | 10.848084 | 2.305686 |
| Animal | 5.887391 | 11.004165 | 3.049897 |
| Animal | 7.0849334 | 12.512583 | 5.792557 |
| Average Retrieval rate | 276.5792/37 = 7.4751 sec | 470.9113/37= 12.7273 sec | 141.6727/37= 2.3876 sec |

**Table 2. Average recognition rate for the overall Face database, Animal database, vehicle database and Flower database.**

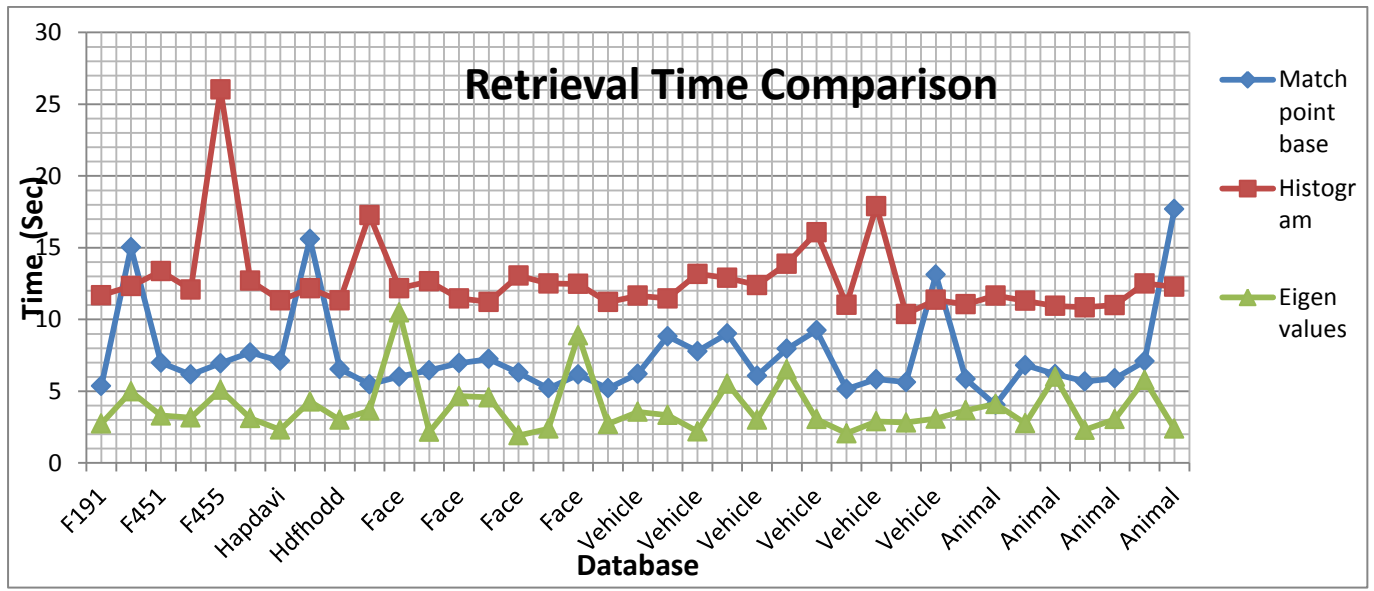

**Figure 4. Graphical representation of retrieval time using three techniques.**

## 5 Conclusion:

This paper proposes three techniques for image retrieval from the various type of database such as human face, vehicle, animal and flower. Three techniques used here are based on match point, color histogram and eigen values. Out of these three techniques, retrieval using eigen value, found to be the best, 2.3876 seconds because only diagonal values of the images are used for comparison because of which retrieval time reduces substantially. Color Image Histogram requires maximum time, 12.7273 seconds because each pixel contributes for the construction of histogram. Lastly required retrieval time of Match Points is in between Eigen values and Histogram Techniques, 7.4751 seconds because only selected match Points contribute for the query image retrieval.

**Future scope:** Limitations of techniques used in this paper are number of pixels contributing the query image retrieval. In histogram technique, every pixel contributes for the plot of Histogram and hence it takes maximum time. Whereas in case of Match Points, selected pixels are used for the image retrieval and in Eigen Values only diagonal elements are used. Therefore retrieval time is least for eigen value technique, maximum for Histogram and for Match Point, retrieval time is in between Eigen Value and Histogram. In order to improve the retrieval time, Wavelets and Multi Resolution Analysis (MRA) can be used. This will result in improvement in directional information and retrieval efficiency. Also can be used identify unknown objects.

## Acknowledgements:

Author would like to thank to a technical paper, Relevance Feedback Techniques in Interactive Content Based Image Retrieval by Yong Rui, Thomas Huang and Sharad Mehrotra, from where my mind trigger with the idea of CBIR which is my research topic. I would like to thank my mentor, my guide Dr. G. K. Kharate, whose guidance could help me to realize the idea of CBIR.

Mohini Sardey is currently serving as HOD and Assistant Professor in AISSMS Institute Of Information Technology, Savitribai Phule Pune University, Pune (Maharashtra State) India. She has more than 20 years of experience in teaching field. She is PhD studnt and doing her research work under the eminent guidance of Dr. G. K. Kharate. She earned her Batchelar degree in Electronics Engineering from Amravati University and M.Tech from Government College Of Engineering, Savitribai Phule Pune University, Pune. Her primary area of research is Image Processing and Machine Vision. She is life member of IETE and ISTE. 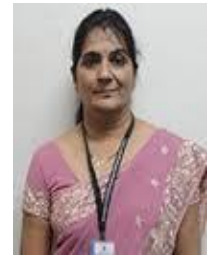

G.K. Kharate is currently Principal, Matoshri College of Engineering and Research Centre, Nashik, Savirtribai Phule Pune University, Pune (Maharashtra), India. A PhD from the University of Pune, Dr Kharate has more than 20 years of teaching experience. He is also a fellow member of the Institution of Electronics and Telecommunication Engineers (IETE) and a life member of many other professional bodies of repute like the Indian Society for Technical Education (ISTE), the Institution of Engineers (India), and the Computer Society of India. Dr. Kharate has also published a number of articles in national and international journals of repute and organized several conferences and workshops in his areas of research. 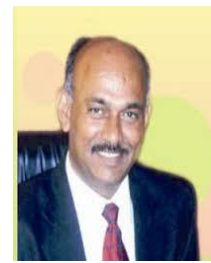